\def\paperlanguage{}
\pdfoutput=1 

\documentclass[letterpaper, 10 pt, conference]{ieeeconf}  

\usepackage{bm}
\usepackage{cite}
\usepackage{flushend}
\include{preamble}

\IEEEoverridecommandlockouts                              

\title{\LARGE \textbf
  {
    \switchlanguage%
    {%
    Design of a Five-Fingered Hand with Full-Fingered Tactile Sensors Using Conductive Filaments and Its Application to \\Bending after Insertion Motion
    }%
    {%
    Design of a Five-Fingered Hand with Full-Fingered Tactile Sensors Using Conductive Filaments and Its Application to Bending after Insertion Motion
    }%
  }
}

\author{Kazuhiro Miyama, Shun Hasegawa, Kento Kawaharazuka, Naoya Yamaguchi, Kei Okada, Masayuki Inaba 
\thanks{*This work was not supported by any organization}
\thanks{K. Miyama,S. Hasegara, K. Kawaharadzuka, N. Yamaguchi, K. Okada, M. Inaba are with JSK Laboratory, Graduate School of Information Science and Technology, The University of Tokyo, 7-3-1 Hongo, Bunkyo-ku, Tokyo 113-8656, Japan
        {\tt\small miyama@jsk.imi.i.u-tokyo.ac.jp}}%
}

\begin{document}

\maketitle
\thispagestyle{empty}
\pagestyle{empty}

\begin{abstract}
  \switchlanguage%
  {%
  The purpose of this study is to construct a contact point estimation system for the both side of a finger, and to realize a motion of bending the finger after inserting the finger into a tool (hereinafter referred to as the bending after insertion motion).
  In order to know the contact points of the full finger including the joints, we propose to fabricate a nerve inclusion flexible epidermis by combining a flexible epidermis and a nerve line made of conductive filaments, and estimate the contact position from the change of resistance of the nerve line.
  A nerve inclusion flexible epidermis attached to a thin fingered robotic hand was combined with a twin-armed robot and tool use experiments were conducted.
  The contact information can be used for tool use, confirming the effectiveness of the proposed method.
  }%
  {%
    本研究では、導電性フィラメントを用いて製作した神経線によって指全周の接触点推定システムを構成し、それを用いた差込後屈曲動作の実現を目的とする。
    関節を含める指全周の接触を知るために、柔軟性を持つ表皮と導電性フィラメントで作られた神経線を組み合わせることで神経内包柔軟表皮を製作し、神経線の抵抗値の変化から推定することを提案する。神経内包柔軟表皮を細い指のロボットハンドに取り付けたものを双腕ロボットと組み合わせ、道具の使用実験を行った。指背面や接触情報を道具の使用に活用することができ、提案する神経内包柔軟表皮の有効性を確認した。
  }%
\end{abstract}

\section{Introduction}\label{sec:introduction}
\subsection{Outline of the Bending after Insertion Motion}
\switchlanguage%
{%
In recent years, robots have been making inroads into society, and there are more and more opportunities for robots to play an active role in the home. Although household support robots that perform a single task are currently active, it is expected that more sophisticated robots will be active in the future. For this purpose, robot hands that can manipulate various tools will be necessary. \par
Tool manipulation with robotic hands has been studied for many years. Some robot hands with many degrees of freedom have thicker fingers due to the incorporation of actuators in the fingers\cite{932538}, which may make it difficult to manipulate tools for insertion. Therefore, we focus on the action of inserting the fingers into the tool and bending them (hereinafter referred to as the bending after insertion motion), with the aim of realizing complex tool use.\par
The bending after insertion motion is adapted to a variety of tool uses such as scissors, drawers, and doorknobs. Although some researches have achieved the use of scissors by attaching them to parallel grippers \cite{parallel-scissors}, they have not been able to achieve the motions required for housework support robots, such as grasping and using tools placed on the plane.\par
In the case of the five-fingered hand, Fukaya et al.\cite{f-hand8229384} compared the conventional hand with the F-hand and found that the F-hand was able to grasp and use 11 types of tools including scissors, but was unable to grip them due to its inflexibility and the conventional hand lacked the degree of freedom to grasp and use scissors. It is expected that the ability to perform finger insertion into tools and environments in addition to common grasping tasks will greatly aid in the imitation of human tasks.\par
Based on the above, the following two elements are considered to be necessary for the post-insertion bending operation.\par

\begin{itemize}
  \item Tactile sensation of the both side of the finger
    \begin{itemize}
    \item After inserting the finger, there is a possibility of contact at any point, whether on the dorsal or palm side of the finger. In order to obtain information on the current contact point and control it, a full-finger tactile sensor, including the joints, is required.
    \end{itemize}
  \item Skeletal structure with thin fingers, imitating the human
    \begin{itemize}
      \item Human imitation hands are important for tool use because they can be driven in a manner similar to human motion. By using a mechanism that imitates the human skeletal structure, it is possible to achieve control that human motion. 
    \end{itemize}    
  \end{itemize}

}%
{%

近年、ロボットの社会進出が進み、家庭でもロボットが活躍する機会が増加している。現在は単一のタスクを行う家事支援ロボットが活躍しているものの、今後はより高機能なロボットが活躍することが期待されており, そのために必要になるのが様々な道具を使用できるロボットハンドである．\par
ロボットハンドによる道具操作は長年研究されている。その中でも単純把持ではない道具の操作は困難であり、1指につき2自由度以上を必要とする複雑な動作では専門化したハンドをタスク毎に作成する場合も多い。
自由度の多いロボットハンドは指の中にアクチュエータを組み込むことで指が太くなっているもの\cite{932538}もあり、差込を行う道具使用が困難になっている場合がある。そのため，指を道具に差し込んで屈曲させる動作(以下、差込後屈曲動作とする)に注目し，複雑な道具使用を実現させることを目的とする．\par
差込後屈曲動作はハサミや引き出し、ドアノブといった多様な道具使用に適応される。一部の研究では並行グリッパに取り付けることでハサミ使用を達成している研究\cite{parallel-scissors}があるものの、床に配置された道具を把持して使用するといった家事支援ロボットに必要とされる動作は達成できていない。\par
これらのことから、差込後屈曲動作動作に必要な要素は、以下の2つが考えられる.
\begin{description}
    \item[Skeletal structure with thin fingers, imitating the human hand]\\
        人体模倣型のハンドは人間の動作と類似した駆動が可能になっているため、道具使用を行う上では重要となっている。人間の骨格構造を模倣した機構にすることで、人間の動作を模倣した制御を実現できる。特に、指を差し込んだ後の屈曲といった動作が重要である。
    \item[指全周の触覚]\\
        差し込んだ後は指の背側、掌側問わずあらゆる場所で接触する可能性がある。現在の接触点の情報を取得して制御を行うためには、関節を含めた指全周の触覚センサが必要になる。
 \end{description}
}%

\subsection{Design and Sensing of the Proposed Method}
\switchlanguage%
{%
\begin{figure}[t]
  \centering
  \includegraphics[width=0.9\columnwidth]{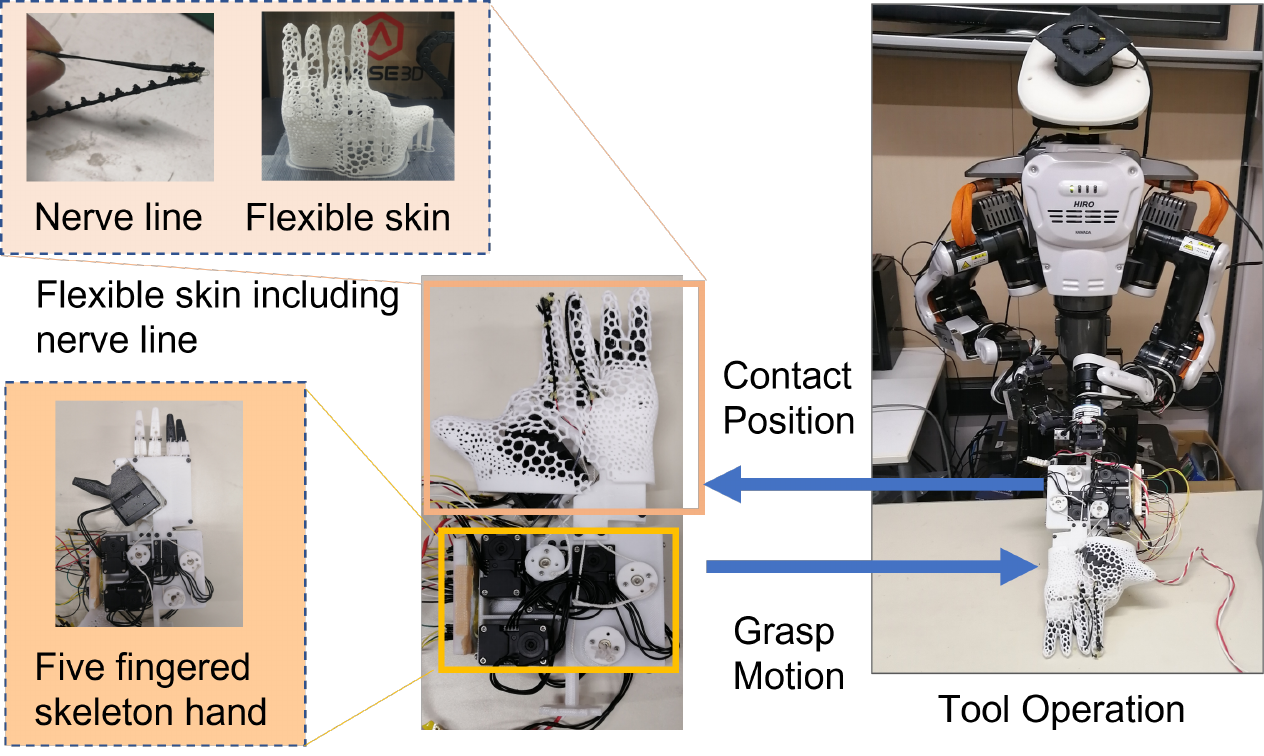}
  \caption{system config of the proposed method}
  \label{fig:skin-config}
\end{figure}
The sensing of the robot hand, such as full-finger tactile sensing, has been studied in a wide range of fields, including grasping and task realization. For grasping, there are methods such as light sensor\cite{koyama2018high}, capacitance sensor\cite{wistort2008electric,escaida20146D}, sound sensor\cite{jiang2012seashell}, etc., which are used for task realization. In humanoids, contact sensors\cite{gerratt2014stretchable} are often used. Most of these sensors are placed on the fingertips and belly of the fingers of the hand, and although they acquire detailed contact and distance information, they rarely acquire information on the back of the fingers and joints.\par
We aim to realize a tool-using task by fabricating a nerve inclusion flexible epidermis capable of detecting contact points of fingers (\figref{fig:skin-config}). The nerve inclusion epidermis consists of a contact sensor made of conductive filaments (hereinafter referred to as "nerve line") and an epidermis made of flexible filaments using a 3D printer, which can be used as a sensor module by covering the robot hand with them. \par
There have been many previous studies on attaching flexible materials such as epidermis to robots, such as a method of fabricating a parallel gripper with a material that deforms to the shape of the object in order to grasp it safely\cite{8830489}, and a method of reducing the impact on the grasped object by combining epidermis and a multi-node skeleton in a three-fingered hand\cite{hirose-san}. For the five-fingered hand, a method that incorporates a contact sensor into the flexible epidermis has been proposed\cite{8594159}, although the joints are not covered.\par
In this study, the epidermis will be used for the safe grasping of objects and for improving frictional force, which have been proposed in the aforementioned studies, as well as for fixing nerve lines. It can be easily integrated into the epidermis, so it can be utilized as a full-3d printed sensing epidermis that can be easily removed from the hand.\par
Finally, we will verify the effectiveness of the robot system developed with the flexible epidermis with nerve line for tool manipulation by experimentally realizing the bending after insertion motion.\par
}%
{%
指全周の触覚のようなロボットハンドのセンシングに関しては、把持やタスク実現などの幅広い分野で研究されている。把持においては光センサ\cite{koyama2018high}や静電容量センサ\cite{wistort2008electric,escaida20146D}、音センサ\cite{jiang2012seashell}などの手法があり、タスク実現を目的としたヒューマノイドにおいては接触センサ\cite{gerratt2014stretchable}などが用いられている。これらのセンサの多くはハンドの指先や指の腹に配置しており、詳細な接触情報や距離情報を取得しているものの、指背面や関節部分の情報を取得することは少ない。\par
本研究では、指全周での接触点を検出可能な神経内包柔軟表皮を製作し、それを指の細い人体模倣ロボットハンドにとるつけたものを双腕ロボットと組み合わせることで、道具使用タスクの実現を目標とする(\figref{fig:skin-config})。神経内包表皮は導電性フィラメントを用いて製作した接触センサ(以下、神経線とする)と柔軟なフィラメントを用いて製作した表皮によって構成されており、これらをロボットハンドに被せることによってセンサモジュールとして使用可能である。\par
表皮のような柔軟素材をロボットに取り付ける先行研究は多く、平行グリッパに関しては近年は物体を安全に把持するために対象物の形状に変形する素材で製作する手法\cite{8830489}や、3指ハンドでは表皮と多節骨格を組み合わせることにより把持物体への影響を減らす手法\cite{hirose-san}などが提案されている。また、五指ハンドでは関節は覆っていないものの柔軟表皮に接触センサを組み込んだ手法\cite{8594159}などが提案されている。
本研究では、表皮を物体の前述の研究でも提唱されていた物体の安全な把持、摩擦力の向上の他に、神経線の固定といった効果を目指して使用する。表皮に簡易に組み込むことができるため、ハンドからの取り外しが容易なfull-3d printedなセンサ付き表皮として活用することができる。\par
最後に、差込後屈曲動作を実験によって実現させることで開発した神経内包柔軟表皮によるロボットシステムが道具操作に効果を有するかを検証する。\par
}%

\section{Contact Point Estimation Module, Nerve Inclusion Flexible Epidermis} \label{sec:nerve-inclusion}
\switchlanguage%
{%
  The nerve inclusion flexible epidermis is an epidermis that can be covered with a robot hand and contains a nerve line as a contact sensor, and is composed of a combination of flexible material filaments and conductive filaments as nerves. \par
  The robot hand is covered with a nerve inclusion flexible epidermis, and the resistance of each nerve line is measured. The module is designed to estimate the contact point by measuring the resistance value that changes.\par
}%
{%
神経内包柔軟表皮とはロボットハンドに被せることのできる、神経線としての接触センサを内包した表皮のことであり、柔軟素材のフィラメントと神経としての導電性フィラメントを組み合わせによって構成されている。神経内包柔軟表皮を取り付けることによって、関節を含む指全周で接触点を検知できるようにすることを目的に設計した。\par
製作したロボットハンドに神経内包柔軟表皮を被せ、各神経線の抵抗値を計測する。接触によって変化する抵抗値を計測すれば接触箇所を推定することが可能になっている。\par
}%

\subsection{Epidermal Flexibility Design} \label{subsec:skin}
\begin{figure}[htbp]
  \centering
  \includegraphics[width=0.9\columnwidth]{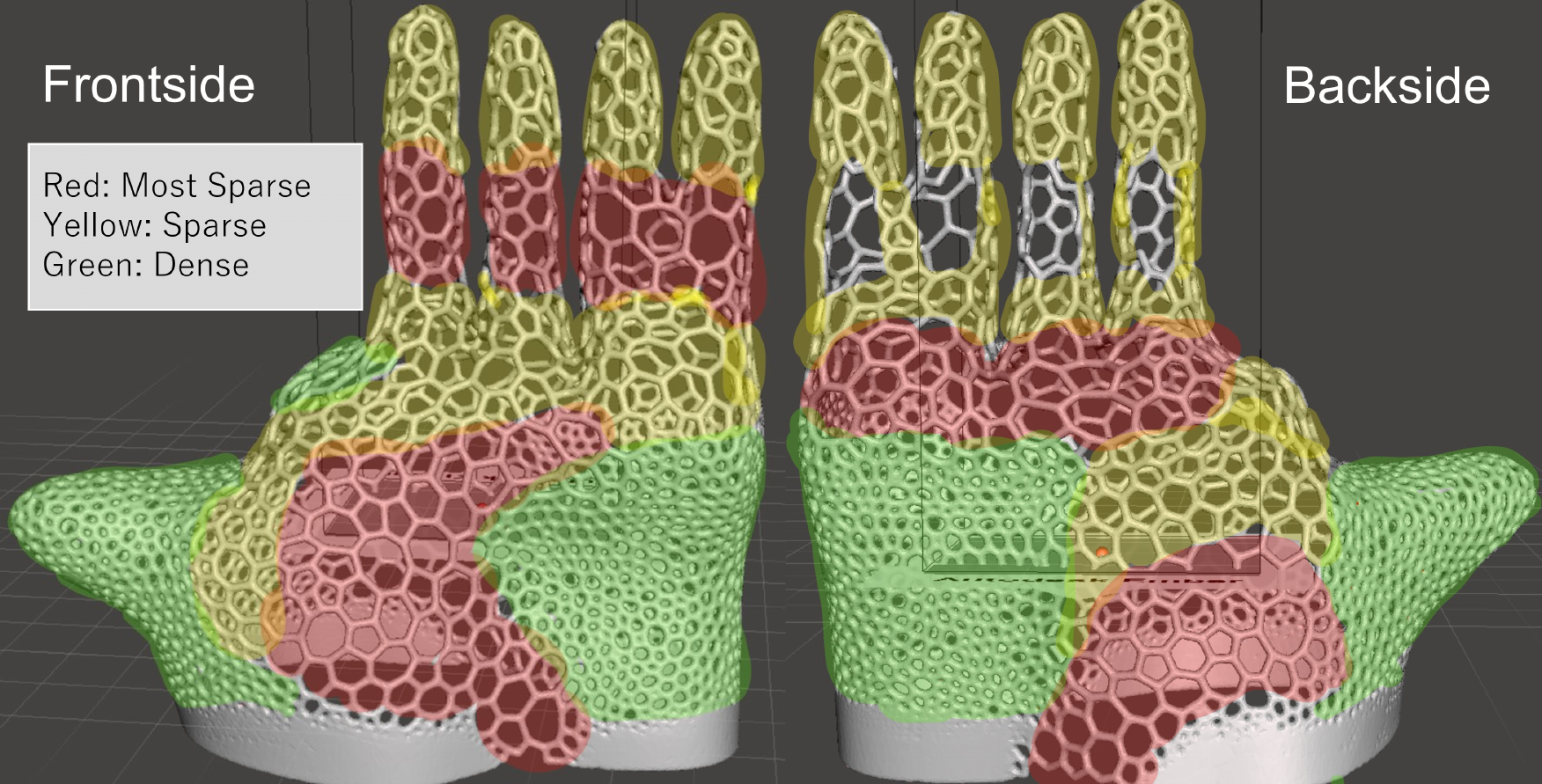}
  \caption{Differences in mesh density settings by part. Low density is used for joints that move a lot, and high density is used for parts that require strength.}
  \label{fig:skin-concept}
\end{figure}

\switchlanguage%
{%
  We used Raise3D Pro2 as s 3D printer, and used TPE filament called TPE60A (produced by HottyPolymer) for making epidermis. In order to obtain flexibility in the epidermis, it is necessary to take some measures for the model of the epidermis. For example, it is possible to make holes in the epidermis only where expansion and contraction are required, or to enable expansion and contraction by adding a bellows structure to the joints. \par
  However, in this study, we used a method of making holes in the epidermis based on the Voronoi structure. The Voronoi structure is a structure that divides the domain of the model according to the roughness of the mesh, and its application to soft machines has been proposed by Debkalpa et al\cite{3DArchtect:Debkalpa:AFM2019}. In their previous study, they claimed that by varying the density of the mesh, it was possible to design a direction in which the hand and fingers would bend easily. In this study, we used their results to design the Voronoi structure of the epidermis model.\par
  The epidermis also has the purpose of acquiring frictional force beyond that of the skeleton. Therefore, it was fabricated by 3D printing using filaments with high frictional force and flexibility. The differences in density design are color-coded in \figref{fig:skin-concept}.
}%
  {%
表皮に柔軟性を獲得するためには、表皮のモデルに対して何らかの対応が必要になる。伸縮が必要な部分のみ穴を開けるといった手法や、蛇腹構造を関節に持たせることで伸縮を可能にするといった手法が考えられるが、本研究では表皮にボロノイ構造に基づいて穴を開けるという手法を用いた。「ボロノイ構造」とは、メッシュの粗さによってモデルの領域を分割する構造であり、Debkalpaらの研究でもソフトマシンへの適用が提案されていた\cite{3DArchtect:Debkalpa:AFM2019}。先行研究においては、メッシュの密度を変化させることでハンドや指が曲がりやすくなる方向を設計できるとしてた。本研究では、その結果を参考にして表皮のモデルのボロノイ構造を設計した。\par
また、表皮には骨格以上の摩擦力を獲得するという目的もある。そのため、摩擦力と柔軟性が高いフィラメントを用いて3Dプリンティングすることで製作した。\par
関節部分は屈曲が必要であるため、低密度にした上で高い延性が必要になる部位には穴を開けた。それに対し指自体や掌は強固な構造にするために高密度に設計したことで、加圧によって低密度の箇所のみが曲がるようになっている。具体的な密度の設計を色分けしたものを\figref{fig:skin-concept}に示す。

}%

\subsection{Neural Line Structure and Contact Point Estimation Method} \label{subsec:nerveline}
\switchlanguage%
{%
The nerve line was made with a conductive TPU filament called Conductive Filaflex(produced by Recreus). The nerve line can be roughly divided into a flat part (A in \figref{fig:line-overview}) and a string-like part (B in \figref{fig:line-overview}), and each parts are insulated. The electrical resistivity of Conductive Filaflex is 3.9[$\Omega\cdot$ cm], so the entire nerve line has a resistance of about 20[$k\Omega$]. \par
\begin{figure}[htbp]
  \centering
  \includegraphics[width=0.8\columnwidth]{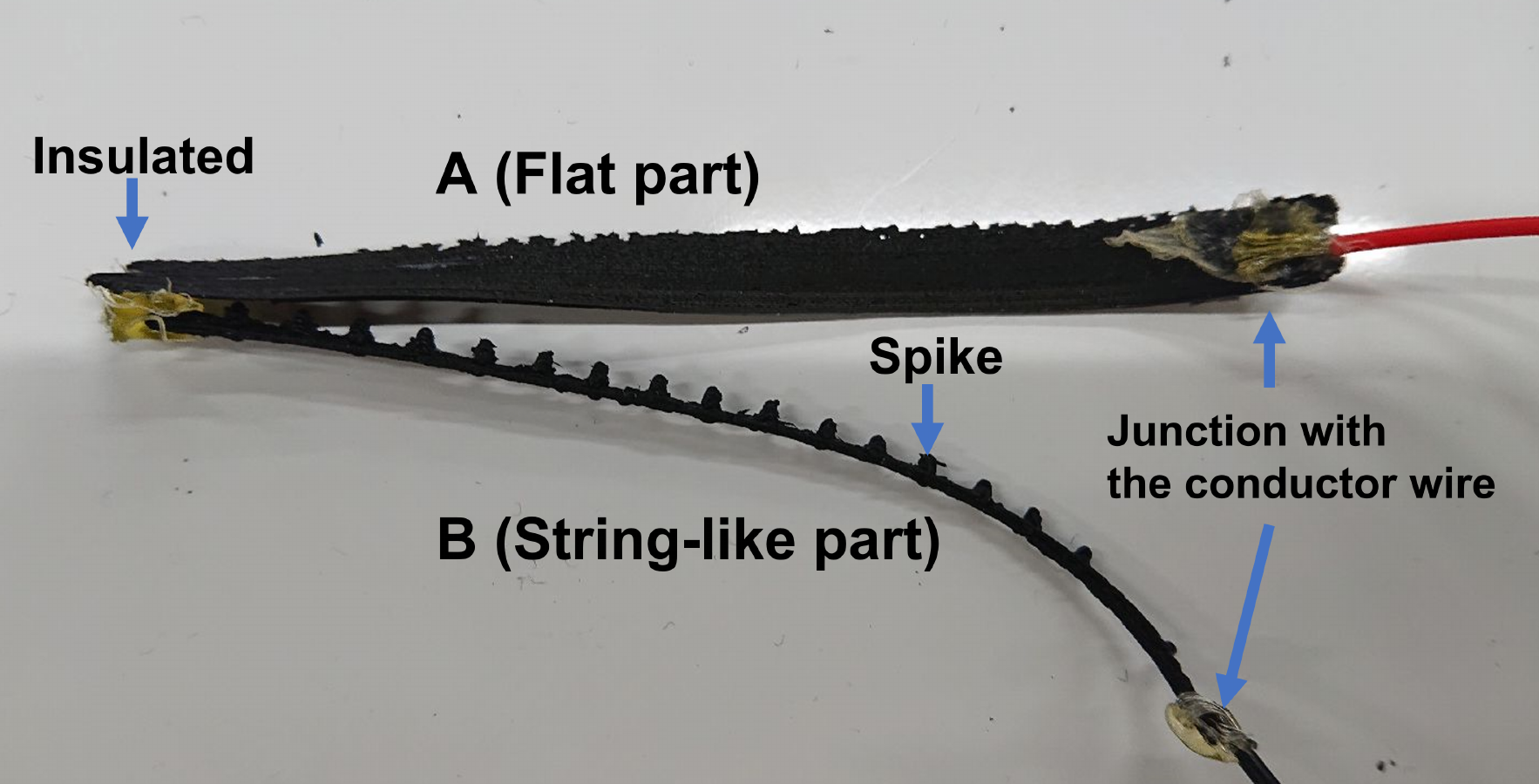}
  \caption{Overview of a nerve line. The two parts are insulated and their opposite sides are the junction with conductor wires}
  \label{fig:line-overview}
\end{figure}

The flat part is 1 mm thick, has a large area, and is flexible and bendable, since it is intended to be placed in contact with the skeleton. The strings-like part have spikes at 5 mm intervals. This is designed to detect contact through the holes in the mesh of the epidermis, since the order of the arrangement is: flat - epidermis - string, as shown in \figref{fig:cross-section}.\par

\begin{figure}[htbp]
  \centering
  \includegraphics[width=0.8\columnwidth]{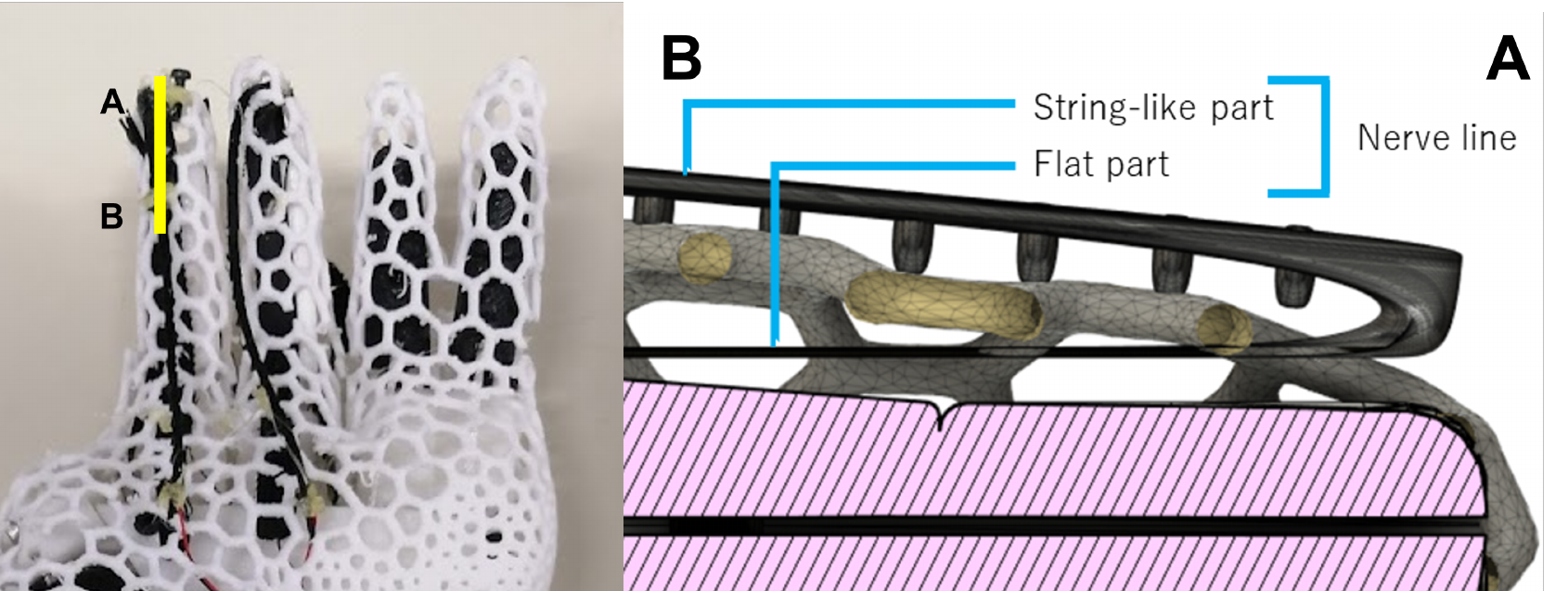}
  \caption{A cross section of the finger. The nerve line is arranged across the epidermis.}
  \label{fig:cross-section}
\end{figure}


In the initial state, the resistance of the nerve line is infinite because both pats are insulated (The reason for insulating both parts is discussed below). Then, when the flat part and the string-like part make contact, the voltage value can be obtained by the conductivity between the two. The farther the contact point is to the connection with the conducting wire, the greater the overall resistance of the nerve line, so the contact point can be estimated by measuring the value of the voltage.\par
Based on the above principles, if the flat and string-like parts are not insulated, the contact at the furthest point from the junction with the conductor wire (The point means fingertip) is almost indistinguishable from the steady state. Since fingertip contact detection is very important in tool manipulation, the two parts are insulated to increase the change in voltage value. \par
However, while insulation has made contact detection more sensitive, the change in voltage values is no longer simply linear. Therefore, we estimated the contact point ratio $p$ using two equations for the fingertip and the other part (\equref{eq:sensor}). $p$ is a value expressed as a percentage of effective total length of sensor used as 1. By defining it this way, it is possible to estimate what percentage of the finger is in contact.\par
$v$ is the sensor data, $V_\mathrm{\, max}$ is the steady state sensor data, $V_\mathrm{\, mid}$ is the sensor data when the nerve line tip is conduction, and $V_\mathrm{\, min}$ is the minimum value of the sensor data. $V_\mathrm{\, max}$, $V_\mathrm{\, mid}$ and $V_\mathrm{\, min}$ are values to be obtained by calibration (Details are given in \secref{subsec:skin-ex}). Ideally, if $p=100$, the finger is not touching anywhere, and below $100$, the finger is touching somewhere. And the closer the value is to $0$, the more contact is occurring at the bottom of the finger.

\begin{equation}
  \label{eq:sensor}
      \begin{split}
          p = 
          \begin{cases}
              100 - \frac{V_\mathrm{\, max} - v}{V_\mathrm{\, max} - V_\mathrm{\, mid}} \times 20 [\%] & (V_\mathrm{\, mid} < v < V_\mathrm{\, max})\\
              \frac{v - V_\mathrm{\, min}}{V_\mathrm{\, mid} - V_\mathrm{\, min}} \times 80 [\%] & (V_\mathrm{\, min} < v < V_\mathrm{\, mid})
          \end{cases}
      \end{split}
\end{equation}

In this robot hand, four nerve wires are used, attached to the dorsal and palmar sides of the index and middle fingers, respectively. It is possible to obtain contact information for the both side of fingers, which are mainly used for tasks. On both sides, the flat parts are placed along the skeleton, and the cord-like parts are pulled out from the gaps in the Voronoi structural mesh of the epidermis of the fingertips.\par
This enables sensing of the area that covers the front and back of the finger, including the joints, and prevents the joints from coming into contact with each other when the finger is bent. Self-contact does not occur in bending because the string-like part is deformed as in \figref{kukkyoku}.

\begin{figure}[htbp]
  \centering
  \includegraphics[width=0.5\columnwidth]{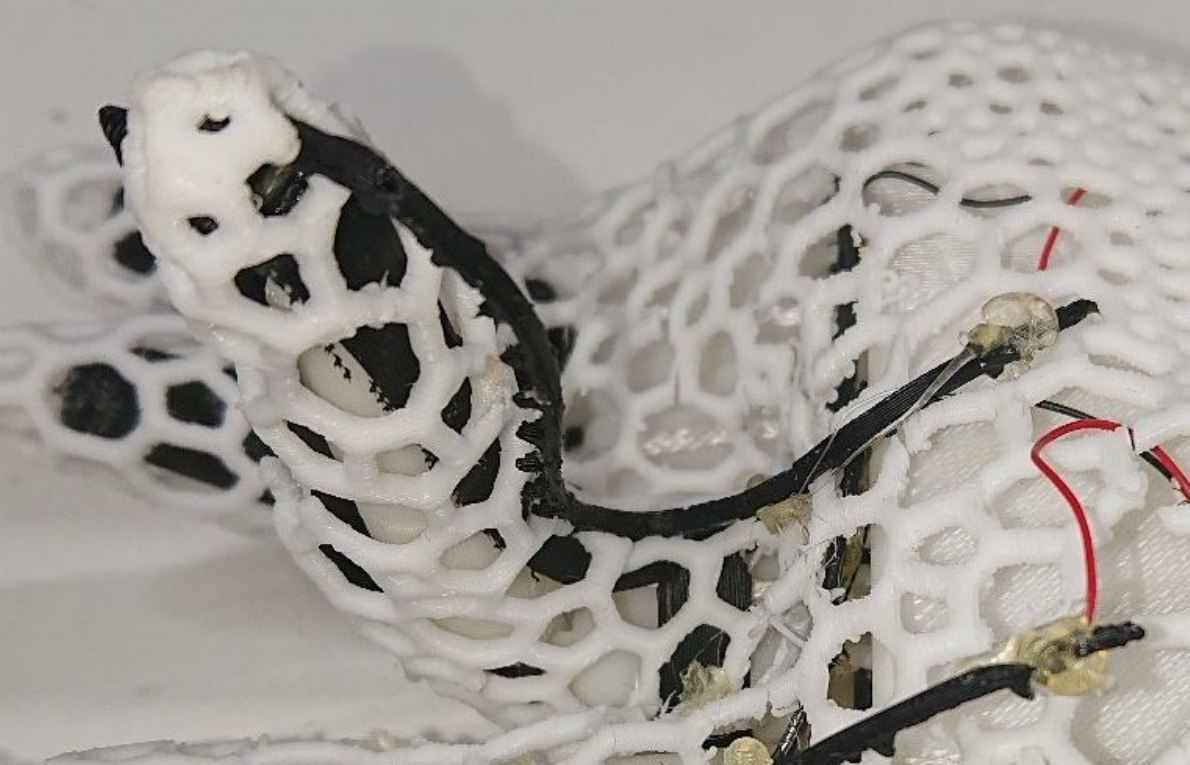}
  \caption{View of the nerve line when the finger is bent. Contact is not detected because the string-like part folds outward.}
  \label{kukkyoku}
\end{figure}

The schematic of a sensor circuit using nerve lines is shown in \figref{fig:cuircuit}. The nerve line is connected to the Arduino and smoothed by applying the RC filter.

\begin{figure}[htbp]
  \centering
  \includegraphics[width=0.9\columnwidth]{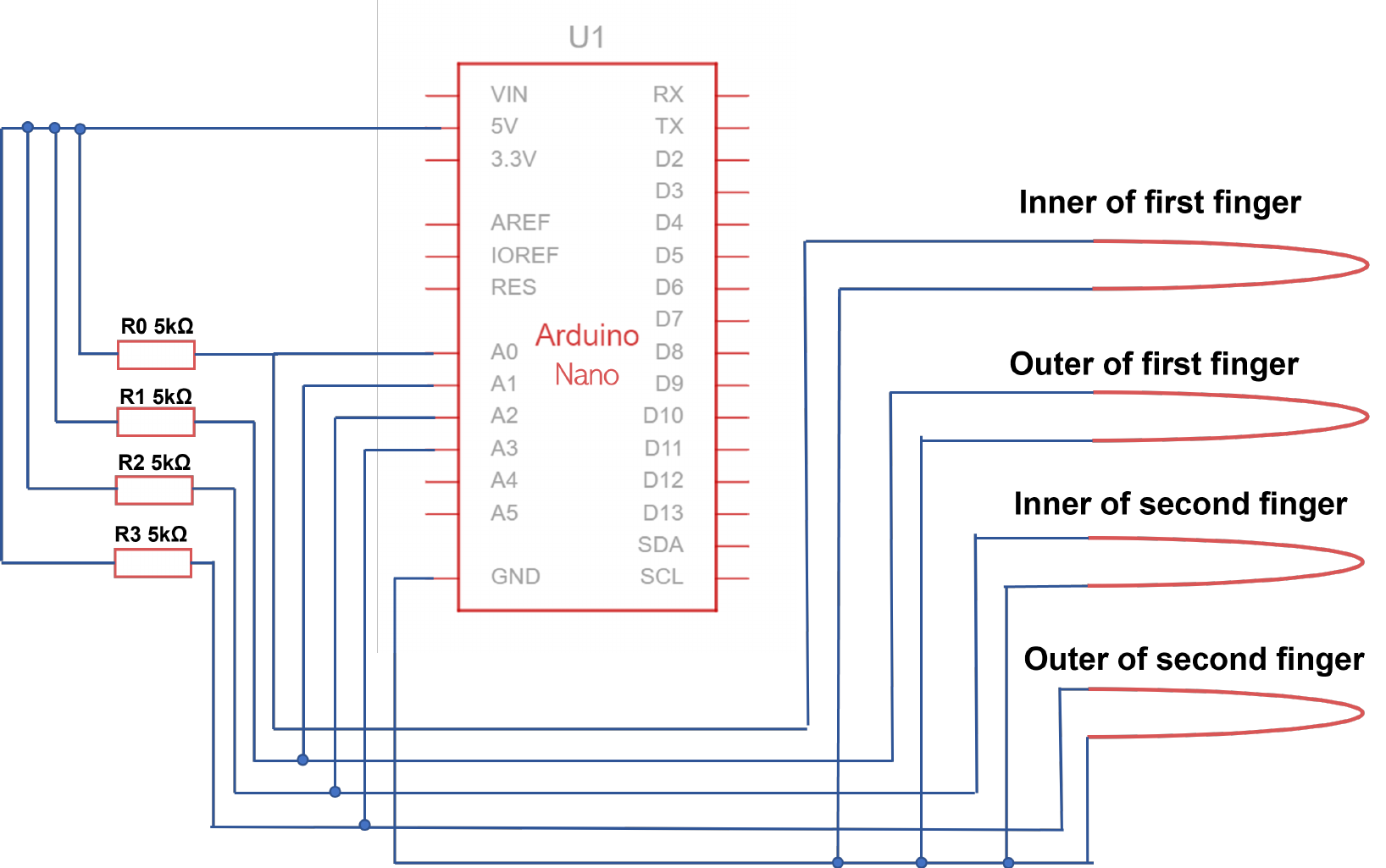}
  \caption{The schematic of a sensor cuircuit. This is consists of nerve line and pull-up resistor.}
  \label{fig:cuircuit}
\end{figure}

}%
{%
神経線は平面部分(\figref{fig:sensor-ex1}左上図のA)と紐状部分(\figref{fig:sensor-ex1}左上図のB)に大別でき、それぞれの接触によってセンサとして機能する。\figref{fig:sensor-ex1}のように接触することで全体の抵抗値が変化し、離すことで即座に定常状態に戻る。平面部分は厚さ1 mmのもので、骨格に接して配置することを目的とするため、広い面積を持ち柔軟に曲がる形状となっている。紐状の部分には5 mm間隔で突起を配置しており、これによって表皮の穴の部分から接触が可能となる。\par

神経線はArduinoに接続されており、Arduino内において\equref{eq:filter}に示すRCフィルタをかけることで平滑化をかけている。ここでは、$v_t$をpublishする値、$v_\mathrm{\, t-1}$を1周期前にpublishした値、$V_\mathrm{\, input}$はArduinoの入力電圧0-5 Vを0-1023のアナログ値に変換したものである。また、$a$は$\frac{1}{1+2\pi dt}$から算出される定数である。\par

\begin{equation}
    \label{eq:filter}
        \begin{split}
            v_\mathrm{\, t} = a \times v_\mathrm{\, t-1} + (1-a) \times V_\mathrm{\, input}
        \end{split}
\end{equation}

神経線は接触点の位置によって抵抗値が線形に変化するため接触センサとして用いることが可能だが、平面状部分と紐状部分の結合部に近づくに連れて抵抗値の変化を検知しづらくなる性質がある。これは、先端で接触した場合は抵抗値の変化が小さく、ノイズとの区別がつかないからである。しかしながら結合部は指の先端に位置し、道具使用において最も感度が必要とされる箇所である。そのため、平面状部分と紐状部分を意図的に切断し、それらをホットグルーで結合することで絶縁し、無限大の抵抗値を持つように設計した。そのため、上部25 ％と下部75 ％で異なった式を用いて接触点割合$p$を推定している(\equref{eq:sensor})。$v$はROS topicで取得したセンサデータ$V_\mathrm{\, max}$は定常状態のセンサデータ、$V_\mathrm{\, mid}$は神経線先端を導通させたときのセンサデータ、そして$V_\mathrm{\, min}$はセンサデータの最小値である。\par

\begin{equation}
    \label{eq:sensor}
        \begin{split}
            p = 
            \begin{cases}
                100 - \frac{V_\mathrm{\, max} - v}{V_\mathrm{\, max} - V_\mathrm{\, mid}} \times 25 [\%] & (V_\mathrm{\, mid} < v < V_\mathrm{\, max})\\
                \frac{v - V_\mathrm{\, min}}{V_\mathrm{\, mid} - V_\mathrm{\, min}} \times 75 [\%] & (V_\mathrm{\, min} < v < V_\mathrm{\, mid})
            \end{cases}
        \end{split}
\end{equation}

これらを柔軟表皮に\tabref{table:sensor}のように取り付けることで、主にタスクに用いる示指と中指の全周の接触情報を取得することができる。

\begin{table}[ht]
    \caption{神経線の取り付け位置}
    \label{table:sensor}
    \centering
    \begin{tabular}{c|c}
        \hline
        sensor ID & 部位\\
        \hline \hline
        0 & 示指掌側\\
		1 & 示指背側\\
        2 & 中指掌側\\
        3 & 中指背側\\
        \hline
    \end{tabular}
\end{table}

理想的には$p=100$ならばどこにも触れていない状態で、$100$以下でどこかに触れている状態となる。そして、$0$に近づくほど指の下の位置で接触が生じていることがわかる。これを用いた行った実験の結果を\figref{fig:sensor-ex1}に示す。この実験では机の上に配置した神経戦を付け根の部分から信号線の接合箇所まで触っていき、どのように電圧が変化するか、そしてそれを元にした接触点割合pの値がどう変わるかを検証した。
\begin{figure}[ht]
    \centering
    \includegraphics[width=0.9\columnwidth]{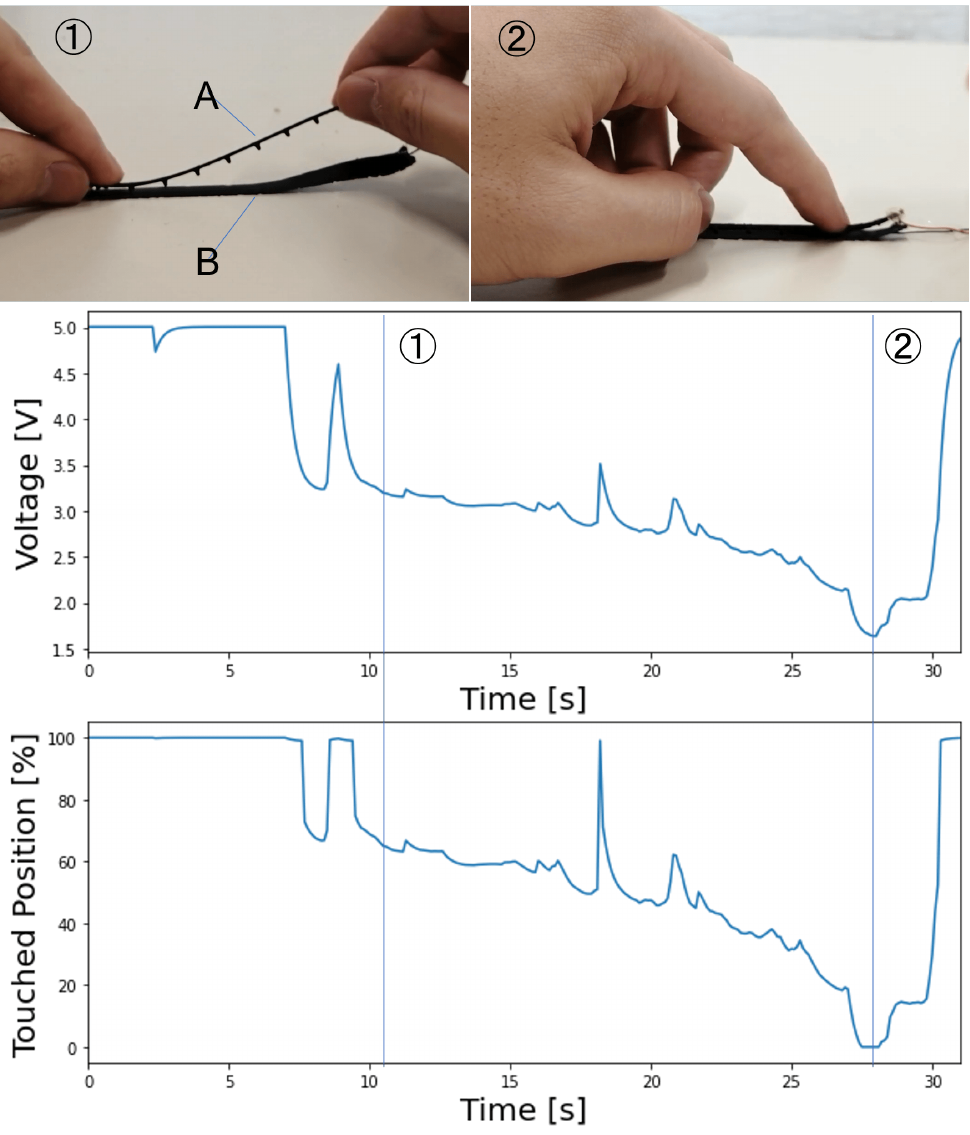}
    \caption{An experiment to verify the effect of nerve line using conductive filaments. Touching from the left side of the figure, the voltage decreases as the finger move to the right side, indicating that the point of contact can be estimated.}
    \label{fig:sensor-ex1}
\end{figure}
神経線は示指、中指それぞれの指に対し2つずつを掌側、背側に配置するような設計になっている。両面とも骨格に沿うように平面部分が配置され、指先の表皮のボロノイ構造メッシュの隙間から紐状の部分が引き出されることで配置されている。これによって、関節を含めた指の前面、背面を覆うような範囲をセンシング可能にしており、指の曲げによって関節部分が接触することを防いでいる。\par

}%

\subsection{Functions of the Nerve Inclusion Flexible Epidermis} \label{subsec:pb-update}
\switchlanguage%
{%
There are two functions to the determination of nerve inclusion flexible epidermis contact: the first is the determination of whether the finger is touching something, and the second is the determination of where on the finger it is touching.
The first starts with the finger not touching anything and changes its behavior when it does. Normally, nerve lines are insulated and the voltage is around 5V, which decreases significantly when contact occurs. 
Therefore, the tip contact can be judged when the contact point ratio $p$ shown by \equref{eq:sensor}is less than the threshold value (set to 90 in the experiments chapter).
This element can be used to judge whether the robot hand has been lowered with the fingers open and made contact with the floor, or whether the robot is able to grasp an object placed on the floor.
The second is the one shown in \equref{eq:sensor}, which is effective when $p$ is less than 80.  In the region where $p$ is less than 80, the resistance value changes linearly, and the contact position can be estimated by averaging over a certain period of time and smoothing out the noise.
Therefore, when the value of $p$ falls below the threshold, it can be determined that the expected position within the 3/4 region of the finger has been touched.
This element can be used to determine where the finger is touching when using scissors, or to determine if the finger is grasping a doorknob, etc. There is a disadvantage in that it cannot determine the point of contact closer to the fingertip when two or more points are touched.
However, for many tasks, the tool is considered stable if it is placed behind the finger, so this is not a problem. 
}%
{%
神経内包柔軟表皮の接触の判定には2つの要素がある。1つ目は指が何かに触れているかの判定、2つ目は指のどの位置に触れているかの判定である。\par
1つ目の接触しているかの判定は、指が何にも触れていない状態から始まり、接触した場合に動作を変化させるものである。通常神経線は絶縁されているため電圧は5V付近であり、接触が発生すると大幅に減少する。そのため、\equref{eq:sensor}で示した接触点割合$p$が閾値(次章の実験では90とした)以下になることで先端の接触を判断することができる。\par
この要素は、指を開いた状態でロボットハンドを下ろしていき床に接触したかの判定や、床に配置された物体を把持できているかの判定に用いることができる。
2つ目のどの位置に物体が接触しているかの判定は、\equref{eq:sensor}で示した$p$が75以下の際に有効になるものである。$p$が75以下の領域では抵抗値が線形に変化し、一定時間の平均をとってノイズを平滑化することで接触位置の推定を行うことができる。そのため、$p$の値が閾値以下になったことで、指の3/4の領域のうちの期待している位置に触れたことを判定することができる。\par
この要素は、ハサミなどを用いる際に指のどこで触れているかの判定や、ドアノブなどを引く際に指の付け根付近でつかめているかの判定に用いることができる。2点以上に触れている際に、より指先に近い接触点は判定できないことが欠点であるものの、多くのタスクでは指の奥に道具が配置されていれば安定状態と判断できるため、問題は少ない。\par

}%

\subsection{Experiment to Verify the Effect of Nerve Inclusion Flexible Epidermis} \label{subsec:skin-ex}
\switchlanguage%
{%
Experiments were conducted to validate a sensor system using nerve inclusion flexible epidermis. Since $V_\mathrm{\, max}$, $V_\mathrm{\, min}$, and $V_\mathrm{\, mid}$ in \eqref{eq:sensor} are different for each nerve line due to differences in 3d printing, contact resistance of conductor connections, etc, calibration is necessary.\par
Before describing the experiment, the calibration method is described. First, steady-state sensor data (voltage divided by 1024) is obtained and the average of 100 times measurements is stored in $V_\mathrm{\, max}$. Contact the point closest to the point where the flat part and the string part are insulated (A in \figref{fig:ex-skin}) and obtain the average. Using the same method, the point closest to the junction with the conductor wire in the range used (B in \figref{fig:ex-skin} if the entire sensor is used) and store the values obtained at each in $V_\mathrm{\, mid}$ and $V_\mathrm{\, min}$.\par
The experimental environment is set at the origin at the junction of the nerve line with the conductor as shown in \figref{fig:ex-skin}. In this experiment, the effective sensor length was set to 80[mm], and the sensor performance was evaluated by comparing the change in the estimated valuee $p$ when the nerve line is contacted at intervals of 5[mm]=6.25[\%]. In addition, to verify the change in performance due to the spikes installed for practicality, the same experiment was conducted on a nerve line without spikes. The results of the two experiments are shown in \figref{fig:ex-skin-result}.\par

\begin{figure}[ht]
  \centering
  \includegraphics[width=0.8\columnwidth]{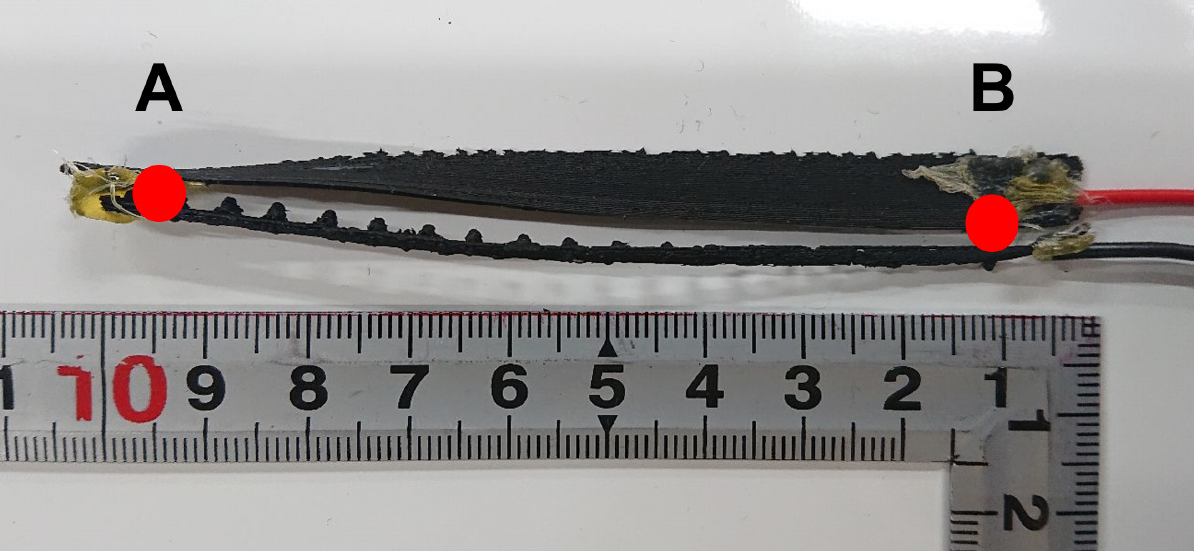}
  \caption{The experimental environment. During calibration, $V_\mathrm{\, mid}$ and $V_\mathrm{\, min}$ can be measured by making contact between the A and B positions.}
  \label{fig:ex-skin}
\end{figure}

\begin{figure}[htbp]
  \centering
  \subfigure[Results in a nerve line with spikes]{
      \includegraphics[clip, width=0.45\columnwidth]{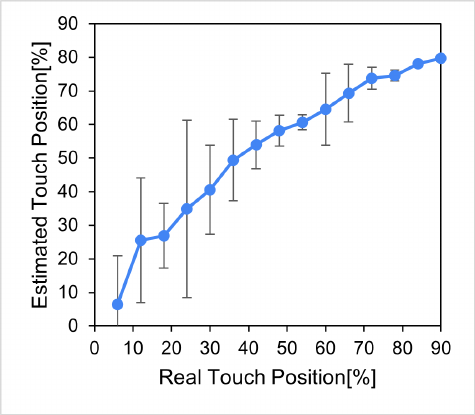}}
  \subfigure[Results in a nerve line without spikes]{
      \includegraphics[clip, width=0.45\columnwidth]{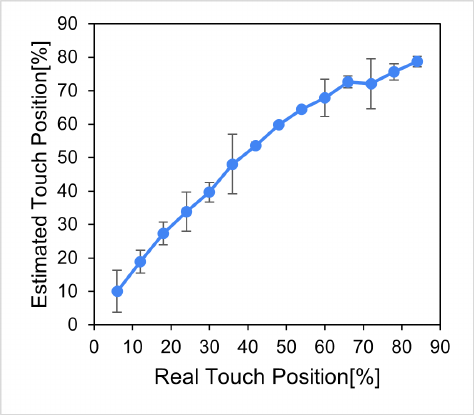}}\\        
\caption{The means and variances for the two experiments are shown. Although the means of both are similar, the variance is larger when there is a spike}
\label{fig:ex-skin-result}
\end{figure}

The experimental results show that the position estimation is linear with and without spikes. When a force was applied in the middle of two spikes on a nerve line with a spike, the estimation results differed significantly depending on which spike made contact. This resulted in a larger variance than in the case without spikes, but the mean was not significantly different. \par
Experimental results confirm that nerve lines can be used to estimate the contact position. 
}%
{%
神経内包柔軟表皮を用いたセンサシステムの検証のための実験を行った。\eqref{eq:sensor}における$V_\mathrm{\, max}$, $V_\mathrm{\, min}$, $V_\mathrm{\, mid}$はセンサの状態によって変化しうるため、センサシステムを用いる前には一定時間抑えている際の平均値を取ることでキャリブレーションをする。\par
その後、\secref{sec:device}で述べるロボットハンドに神経内包柔軟表皮を取り付け、示指と中指の手の平側を指先から付け根にかけて触れていく。作成したハンドモデルに、接触位置の推定によって色を変化させるマーカーを組み合わせることで接触点推定が適切に行われていることを確認した(\figref{fig:sensor-ex1})。実験の際の結果を\figref{fig:sensor-graph}に示す。

神経線による電圧は定常状態では5 Vだが、示指先端の接触を検知すると4 V程度まで低下する(\figref{fig:sensor-graph}-1)。続いて付け根に指を近づけていくが、電圧値は最小で3.5 V程度と大きくは変化しない。しかしながら、電圧から算出された接触点割合Pは60\%から0 \%まで低下することで接触点の変化を推定できていることが確認できる。

}%

\section{Five-Fingered Robot Hand} \label{sec:device}
\switchlanguage%
{%
  The specifications of the robot hand developed in this study were determined with the goal of making the diameter of the fingers close to that of a human being in order to perform insertion motions into human tools.\par
  According to a survey using a method developed by AIRC\cite{kuo2020developing}, the mixed gender average of the joint width of the index finger in humans is 17.2 mm in thick part and 14.9 mm in thin part. Therefore, we designed the joint width to be 9 mm at the thinnest point and 14 mm at the thickest point in order to achieve the same level with the skin attached. \par
  In order to achieve the same level of flexibility after insertion, we aimed to have fingers with multiple degrees of freedom, which we pointed out as necessary elements in \secref{sec:introduction}.
  The main fingers used are the thumb, index and middle fingers, and it is important to realize dorsal abduction and palmar adduction of these fingers. In order to drive them in a design that maintains the slender fingers, the posture is controlled by changing the wire length with seven actuators. The reason why actuators instead of springs are used to drive the extension side is that springs do not have enough output to use drawers or open and close scissors. The actuator and wire arrangement is shown in \figref{fig:device-overview}.\par
  \begin{figure}[ht]
    \centering
    \includegraphics[width=0.9\columnwidth]{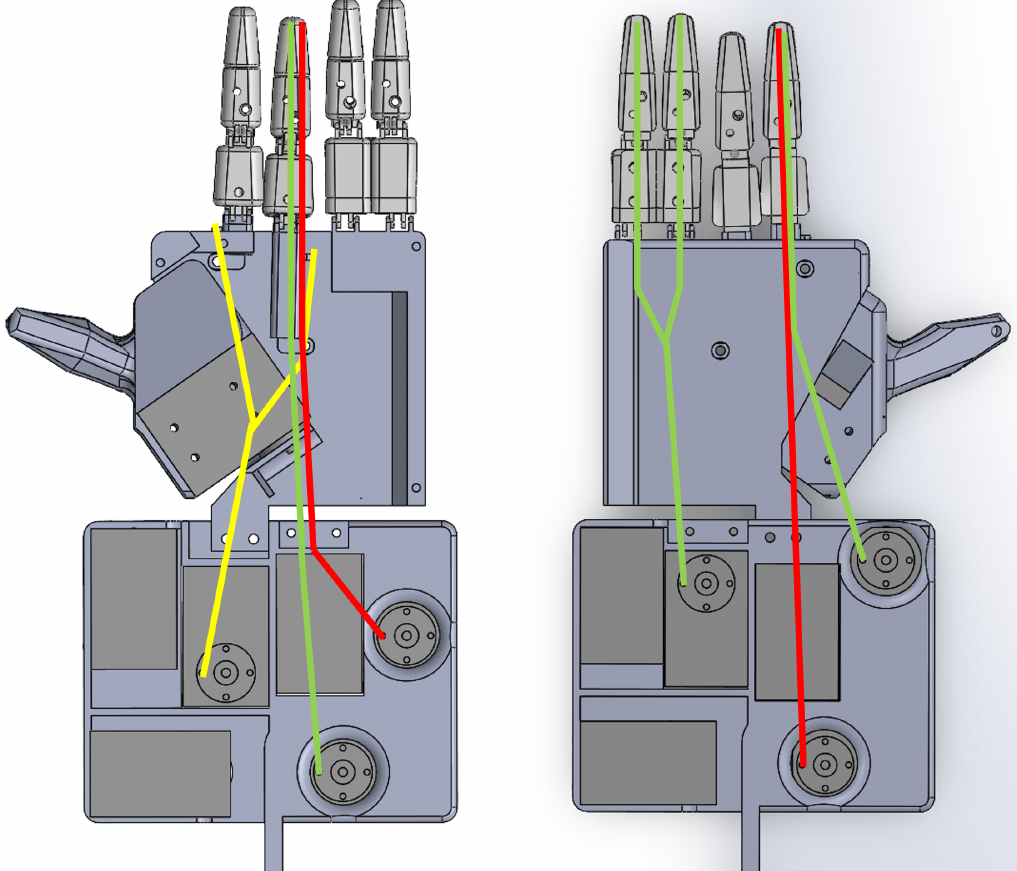}
    \caption{Overall view of the developed device and the functions of the seven actuators. Color coded according to the role of the wire.}
    \label{fig:device-overview}
\end{figure}
  Red lines in the figure represent the wire for extension, and green lines represent the wire for bending. The yellow wire is the wire that operates the internal rotation mechanism. The internal rotation mechanism is used to hold the fingers in place after inserting them into the tool, and an example of it in operation when using scissors is shown in \figref{inserting}.\par
  \begin{figure}[ht]
    \centering
    \includegraphics[width=0.9\columnwidth]{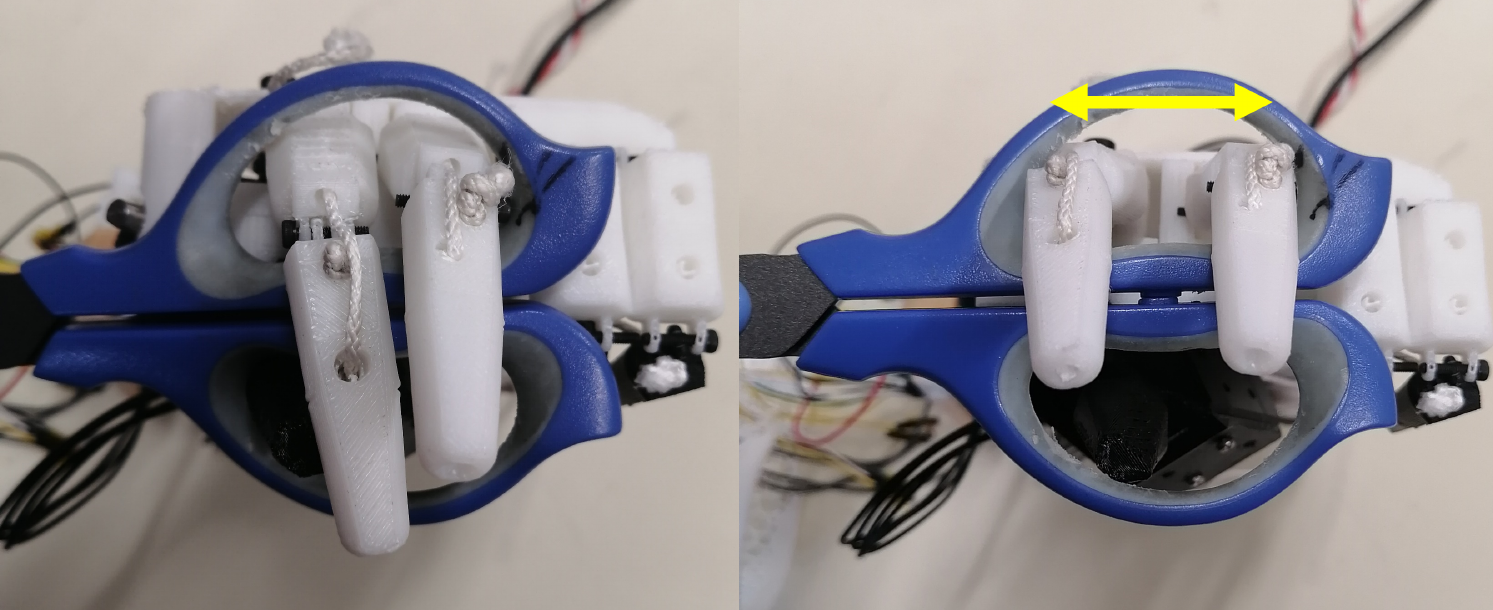}
    \caption{Example of utilizing the internal rotation mechanism. Combined with finger insertion, the tool is fixed in place.}
    \label{inserting}
\end{figure}
  The relationship between the displacement of the wire length $x$ mm and the rotation angle $\theta$ rad can be expressed as \equref{eq: eq1} using the pulley diameter $r$ mm. The optimal displacement of the servo motor other than the one for the rotation of the thumb is non-trivial. Therefore, the optimal displacement of the servo motor was calculated by marking the wire and pulling it at the normal and maximum displacements in the normal operation, and used for the control. \par
  }%
{%
jojoj
本研究で開発したロボットハンドは、人間用の道具への差込動作を行うために、指の直径を人間のものに近づけることを目標に仕様を決定した。AIRCの開発した手法による調査\cite{kuo2020developing}によると、人間の示指の関節幅の男女混合平均は近位では17.2 mm, 遠位では14.9 mmとなっているため、表皮を装着した状態で同程度になるために最も細い部分で9 mm、最も太い部分で14 mmになるよう設計した。2章で述べた必要な要素として指摘した、差し込んだ後の屈曲などを実現するために他自由度の指を有することに目標とした。\par
主に用いるのは母指、示指、中指であり、これらの背側外転と掌側内転を実現することが重要である。細い指を維持したままそれらを駆動させるために、7つのアクチュエータによってワイヤ長を変化させることで姿勢を制御する。ワイヤの長さの変位$x$ mmと回転角$\theta$ radの関係はプーリー径$r$ mmを用いることで\equref{eq: eq1}のように表せる。母指の回転のためのサーボモータ以外のサーボモータの最適な変位は自明ではない。そのため、通常動作における正常時と最大変位時にワイヤに印をつけ実際に引くことで計測的に算出し, 制御に用いた. \par
}%
\begin{equation}
	x = r\theta
	\label{eq: eq1}
\end{equation}

\section{Tool Manipulation Experiment} \label{sec:exp}
\switchlanguage%
{%
Using tools with a five-fingered skeletal robot hand combined with nerve inclusion flexible epidermis, the effect of the control using nerve inclusion flexible epidermis is verified. 
In this study, grasping and manipulation of scissors is performed as a representative example of the bending after insertion motion, and picking up a pen is performed to show that the robot hand can adapt to tools other than the post-insertion bending motion.
Tool use experiments will be conducted on two tools with different motions, and the results will be presented.

}%
{%
神経内包柔軟表皮を組み合わせた五指骨格ロボットハンドによる道具使用を行うことで、神経内包柔軟表皮を用いた制御の効果を検証する。本研究では差込後屈曲動作の代表的な例として引き出しとハサミの把持、操作を行い、差込後屈曲動作以外の道具にも適応可能であることを示すためペンの把持を行う。異なった動作を行う3つの道具に関しての道具使用実験を行い、その結果を示す。\par
ロボットハンドを左手に取りつけたHIRONXの前に机を配置し、head cameraに机全体が収まるように姿勢を変更する。Bounding boxのsubscribeを行って予め道具の位置を取得し、その位置を記録した状態から実験を開始する。\par

}%

\subsection{Scissors Holding Experiment} \label{subsec:p-s-exp}
\begin{figure}[ht]
  \centering
  \includegraphics[width=0.8\columnwidth]{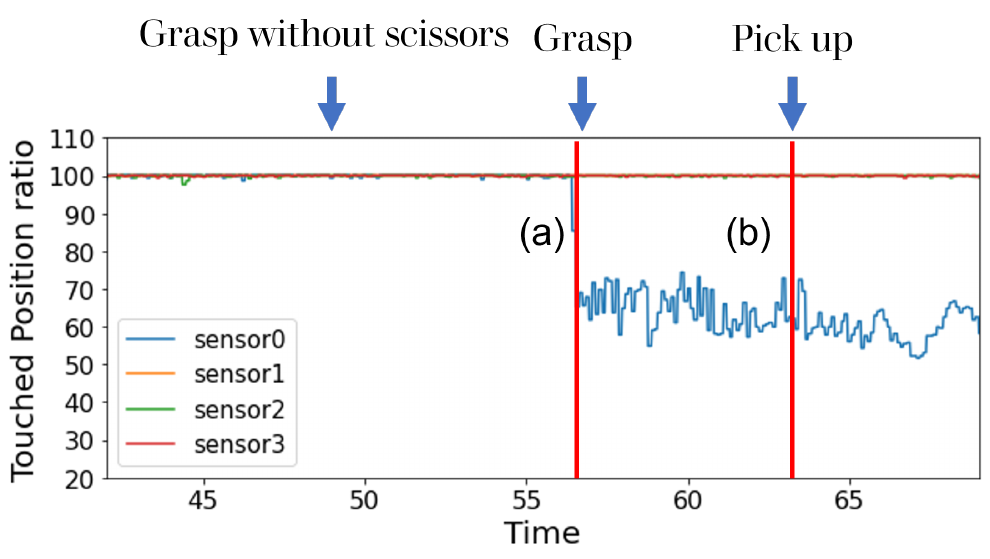}
  \caption{Contact point ratio P when grasping the scissors}
  \label{fig:pickup-graph}
\end{figure}

\begin{figure}[htbp]
  \centering
  \subfigure[Inserting fingers to the hole of scissors]{
      \includegraphics[clip, width=0.4\columnwidth]{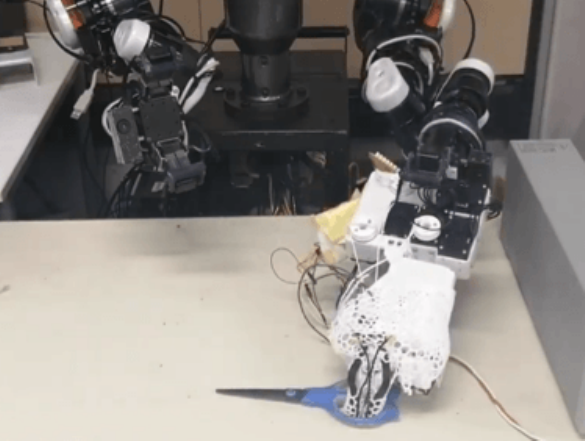}}
  \subfigure[Picking up scissors]{
      \includegraphics[clip, width=0.4\columnwidth]{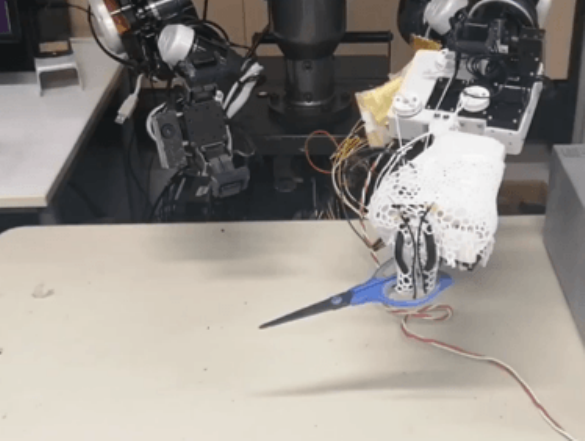}}\\        
\caption{Figure of scissors grasping experiment using nerve inclusion flexible epidermis.}
\label{fig:pickup-scissors}
\end{figure}

\switchlanguage%
{%
The use of scissors is realized by using the nerve inclusion flexible epidermis as a sensor. The position of the scissors is obtained using image recognition with HSI color filter using the information from the head camera, and the robot hand is moved directly above the scissors.
After that, the fingers are lowered to the specified height, and the grasping posture is set by closing the fingers. The neural line detects the contact based on the value of the contact point fraction $p$ in \equref{eq:sensor}, and if it exceeds the threshold, it is considered to have grasped. The threshold value in this experiment was set to $p=90$.\par
\figref{fig:pickup-graph} shows the graph of the sensor data when lifting the scissors. In this experiment, after acquiring the position of the scissors by image recognition, we intentionally moved the scissors to compare the behavior with the failure of grasp. When the grasping was performed without the scissors in place, there was no significant change in the sensor data (\figref{fig:grasp-graph}-Grasp without scissors).
This is because no neural line contact occurs in grasping when the object is not grasped. In this case, since the value of $p$ is not below the threshold, the initial posture is returned and the hand is lifted to the pre-grasp posture.
Then, the person returns the scissors to the previous position, and the scissors are grasped by the robot hand (\figref{fig:pickup-scissors}-(a)). Here, there was a change in the grasping posture, but there was also a significant change in the sensor data.\par
Since this is below the threshold value of $p=90$, we judged that the grasp was successful and lifted the finger in the grasp position (\figref{fig:pickup-scissors}-(b)).
}%
{%
次に、神経内包柔軟表皮をセンサとして用いることでハサミの使用を実現する。ハサミの位置をhead cameraの情報を用いたHSIカラーフィルタによる画像認識を用いて取得し、ロボットハンドをハサミの真上に移動させる。その後指を指定された高さまで下ろしていき、指を閉じることで把持姿勢とする. \equref{eq:sensor}の接触点割合$p$の値から神経線が接触を検知し、閾値を超えた場合は掴んだと判断する。この実験の際の閾値は$p=90$とした。\par
\figref{fig:pickup-graph}にハサミを持ち上げる際のセンサデータのグラフを示した。今回の実験では画像認識でハサミの位置を取得した後で、意図的にハサミを移動させて把持失敗時との動作の比較を行った。ハサミを配置しない状態で把持を行った場合、センサデータに大きな変化はなかった(\figref{fig:grasp-graph}-Grasp without scissors)。これは、物体を挟まない場合の把持では神経線の接触が発生しないためである。この際、$p$の値が閾値を下回らないため初期姿勢に戻り、ハンドを持ち上げることで把持前の姿勢となる。\par
続いて、ハサミを以前の位置に戻すことでハサミの把持を行う(\figref{fig:pickup-scissors}-(a))。ここで把持姿勢に変化したが、センサデータにも大きな変化があった。示指の掌側の神経線(sensor0)のセンサ$p$の値が70~80に低下していることがわかる。これは閾値としていた$p=90$を下回るため、把持が成功したと判断して把持姿勢のまま持ち上げを行う(\figref{fig:pickup-scissors}-(b))。\\
}%

\subsection{Scissors Operating Experiment} \label{subsec:u-s-exp}
\begin{figure}[htbp]
  \centering
  \includegraphics[width=0.8\columnwidth]{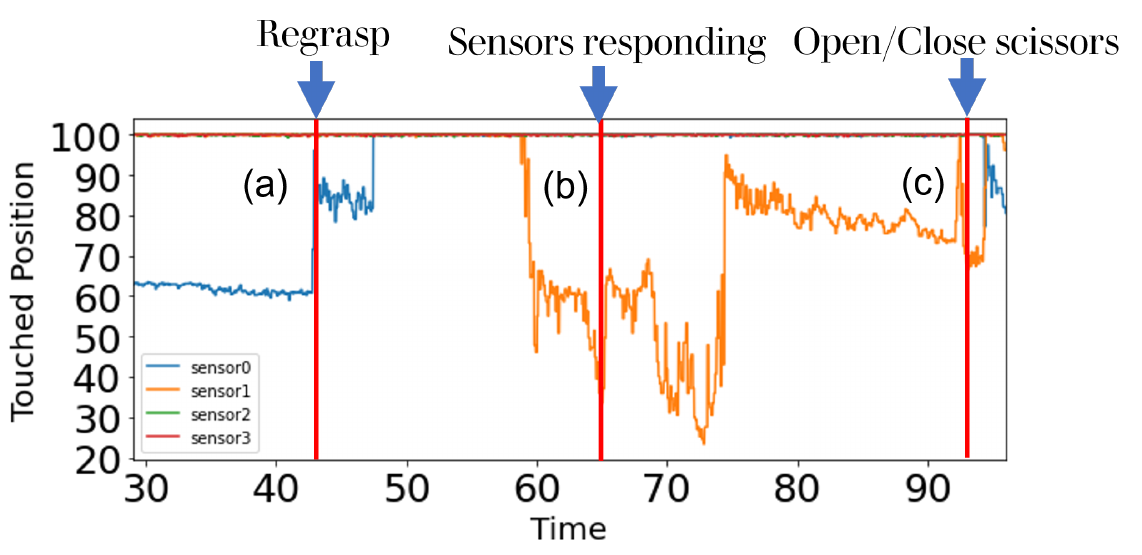}
  \caption{Contact point ratio P when the scissors are regrasped and used.}
  \label{fig:grasp-graph}
\end{figure}

\begin{figure}[htbp]
  \centering
  \subfigure[Initial posture after picking up scissors]{
      \includegraphics[clip, width=0.45\columnwidth]{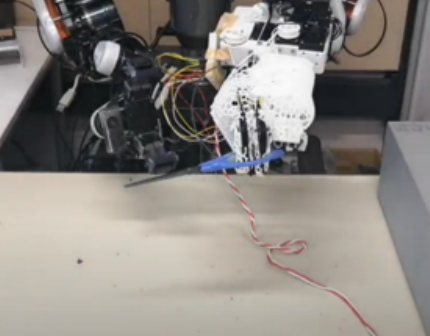}}      
  \subfigure[Grabbing the scissors with the right hand.]{
      \includegraphics[clip, width=0.45\columnwidth]{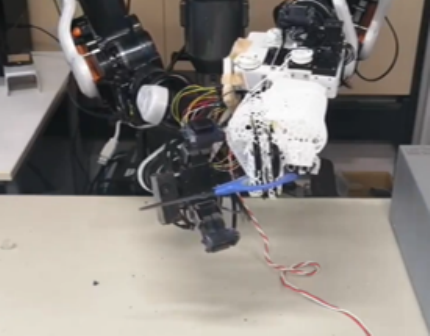}}\\    
  \subfigure[Liftting up the scissors until they move to the base of the finger.]{
      \includegraphics[clip, width=0.45\columnwidth]{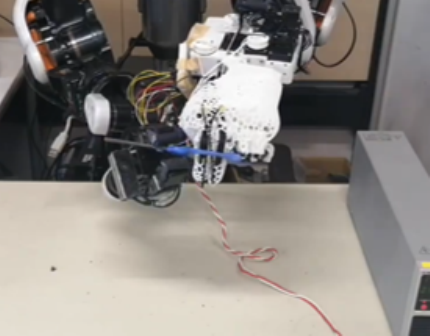}}
\subfigure[Using scissors]{
  \includegraphics[clip, width=0.45\columnwidth]{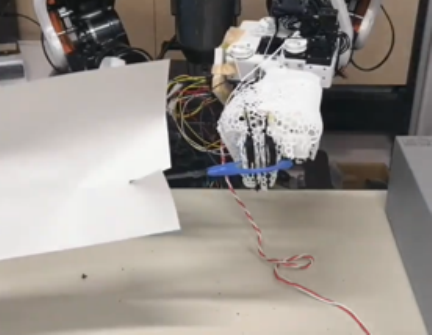}}
\caption{Figure of scissors operating experiment using nerve inclusion flexible epidermis.}
\label{fig:grasp-scissors}
\end{figure}

\begin{figure}[htbp]
  \centering
  \includegraphics[width=0.8\columnwidth]{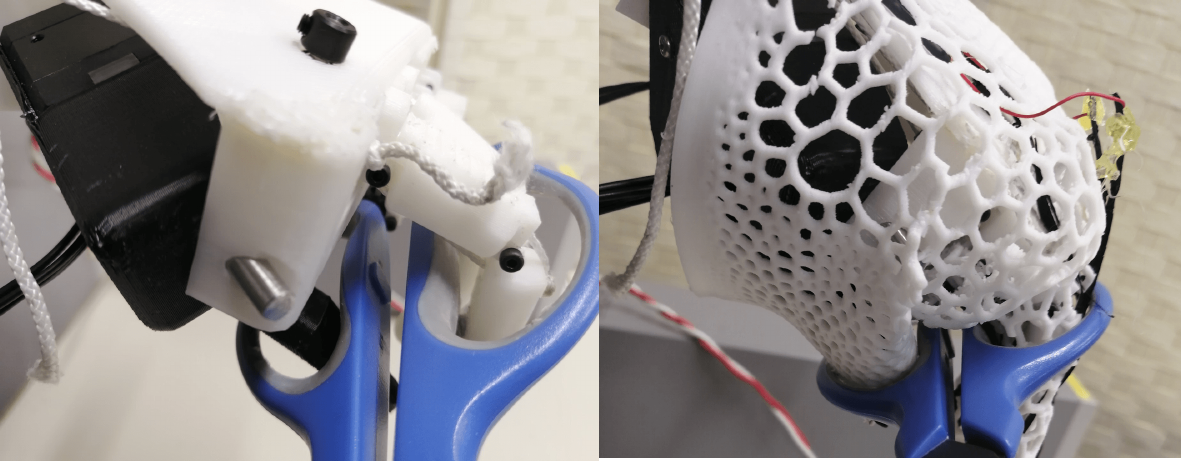}
  \caption{The scissors are fixed in place by inserting and bending the fingers, as in grasping with the skeleton alone.}
  \label{fig:holdpose}
\end{figure}

\switchlanguage%
{%
In order to operate the scissors, the fingers must be closed with moving scissors to the base of fingers.
Change the left and right arms that lifted the scissors to the specified posture and grasp the threaded part of the scissors with the gripper of the right hand (\figref{fig:grasp-scissors}-(a)).\par
After confirming the grasp, change the hand to the pre-grasp state and rotate the right wrist so that it rotates 20° around the x-axis. Then, each time the right hand is moved by 5 mm, the position of the scissors on the fingers is determined. In this case, too, a threshold value is predetermined according to the target position, and when the value falls below the threshold value, it is judged that the scissors have been moved to the back of the finger and the movement is stopped.
Here, sensor1 (the nerve line at the back of the index finger) responded and fell below the threshold $p=50$ for the contact position at \figref{fig:grasp-scissors}-(b) (\figref{fig:grasp-graph}-sensor responding).\par
Detecting that it was possible to move the scissors to the base of the finger by dropping below the threshold, the robot hand changed to the grasping posture and grasped the scissors.
\figref{fig:holdpose} shows the robot hand with only the skeleton and with the nerve inclusion flexible epidermis while grasping the scissors. It can be seen that the scissors are fixed when the index and middle fingers are bent, and the post-insertion bending motion is achieved. When the mother finger was driven in this state, the scissors could be manipulated (\figref{fig:grasp-scissors}-(c)).
}%
{%
ハサミの操作を行うためには、ハサミを指の奥に押し込んだ状態で指を閉じる必要がある。ハサミを持ち上げた左腕と右腕を指定された姿勢に変更し、ハサミのネジ部を右手のグリッパで把持する(\figref{fig:grasp-scissors}-(a))。把持を確認したらハンドを把持前状態に変更し、x軸を中心に20°回転させるように右手首を回転させる。
その後右手を5mm移動するたびに指のどの位置にハサミが当たっているかを判断する。この場合も目標とする位置によって閾値を予め決めておき、閾値以下になった場合に指の奥にハサミを移動させることができたと判断して動きを止めるようにする。ここで、sensor1(示指の背面の神経線)が反応し、\figref{fig:grasp-scissors}-(b)の際に接触位置の閾値$p=50$を下回った(\figref{fig:grasp-graph}-sensor responding)。
閾値を下回ったことで指の奥にハサミを移動させることができたことを検知し、把持姿勢に変更してハサミを掴んだ(\figref{fig:grasp-scissors}-(c))。\par
\figref{fig:kukkyoku}はハサミを把持している際のロボットハンドの骨格のみの場合と神経内包柔軟表皮を装着した際のものである。これを見ると示指と中指が折り曲げられた状態ではハサミは固定されていることがわかり、差込後屈曲動作が達成されていることが確認できる。この状態で母指を駆動すると、ハサミの操作が行えた(\figref{fig:grasp-scissors}-(c))。
}%

\section{Conclusion} \label{sec:conclusion}
\switchlanguage%
{%
In this study, our goal was to develop a tool manipulation system using a nerve inclusion flexible epidermis.
We focused on the bending after insertion motion, in which a finger is inserted into a tool and then bent, and confirmed that a skeletal structure with thin fingers imitating those of the human body and a tactile sense of the both side of the finger were necessary to realize this motion. The nerve inclusion flexible epidermis developed for the tactile sensation of the entire finger was fabricated by 3D printing with flexible materials, and consists of an epidermis whose flexibility can be designed by Voronoi structure and nerve wires that can detect contact by using conductive filaments. We proposed a method to estimate the contact point of the hand by combining it with a multi-degree-of-freedom robot hand that is driven by wires to make the fingers thinner.\par
 By integrating the above robot hand and HIRONX, we confirmed that the robot can cooperate with other robots and showed its scalability.
 We verified the effect of sensing by the neural envelope flexible epidermis, and confirmed that the sensor data can be feed back to the hand control.
 From these results, we were able to show that our proposed method is effective in tool manipulation, including bending after insertion motion.
}%
{%
本研究では、神経内包柔軟表皮を用いた制御による道具操作を目標としていた。道具操作の中でも道具に指を差し込んだ後で屈曲させる差込後屈曲動作に着目し、実現のためには人体模倣の細い指を有する骨格構造、指全周の触覚が必要であることを確認した。指全周の触覚のために開発した神経内包柔軟表皮は柔軟素材による3Dプリントで製作しており、ボロノイ構造によって柔軟性を設計できる表皮と、導電性フィラメントを用いることで接触を検知可能な神経線で構成されている。それを指を細くするためにワイヤ駆動とした多自由度のロボットハンドと組み合わせ、ハンドの接触点推定を行える手法を提案した。以上の製作したロボットハンドとHIRONXを統合したことで他のロボットとの協調動作が可能であることを確認し、拡張性を示した。\par
また、開発したロボットハンド用いてハサミ、引き出し、ペンの3種の道具使用実験を行い、着目した要素が差込後屈曲動作に有効であることを示した。神経内包柔軟表皮によるセンシングの効果を検証し、センサデータを用いることでハンドの制御にフィードバック可能であることを確認した。\par
これらから、差込後屈曲動作を始めとする道具操作において、本研究の提案手法が有効であることを示すことができた。\par
}%

{
  \bibliographystyle{IEEEtran}
  \bibliography{main}

\begin{thebibliography}{10}
\providecommand{\url}[1]{#1}
\csname url@rmstyle\endcsname
\providecommand{\newblock}{\relax}
\providecommand{\bibinfo}[2]{#2}
\providecommand\BIBentrySTDinterwordspacing{\spaceskip=0pt\relax}
\providecommand\BIBentryALTinterwordstretchfactor{4}
\providecommand\BIBentryALTinterwordspacing{\spaceskip=\fontdimen2\font plus
\BIBentryALTinterwordstretchfactor\fontdimen3\font minus
  \fontdimen4\font\relax}
\providecommand\BIBforeignlanguage[2]{{%
\expandafter\ifx\csname l@#1\endcsname\relax
\typeout{** WARNING: IEEEtran.bst: No hyphenation pattern has been}%
\typeout{** loaded for the language `#1'. Using the pattern for}%
\typeout{** the default language instead.}%
\else
\language=\csname l@#1\endcsname
\fi
#2}}

\bibitem{932538}
J.~Butterfass, M.~Grebenstein, H.~Liu, and G.~Hirzinger, ``Dlr-hand ii: next
  generation of a dextrous robot hand,'' in \emph{2001 IEEE International
  Conference on Robotics and Automation}, vol.~1, 2001, pp. 109--114.

\bibitem{parallel-scissors}
T.~Akaike, Y.~Tsumaki, R.~Tadakuma, and M.~Yamano, ``A parallel gripper with
  capability to use various tools,'' in \emph{2012 SICE Annual Conference},
  2012, pp. 2139--2143.

\bibitem{f-hand8229384}
N.~Fukaya and Y.~Ogasawara, ``Development of humanoid hand with cover
  integrated link mechanism for daily life work,'' in \emph{2017 IEEE 6th
  Global Conference on Consumer Electronics}, 2017, pp. 1--4.

\bibitem{koyama2018high}
K.~{Koyama}, M.~{Shimojo}, T.~{Senoo}, and M.~{Ishikawa}, ``High-speed
  high-precision proximity sensor for detection of tilt, distance, and
  contact,'' \emph{IEEE Robotics and Automation Letters}, vol.~3, no.~4, pp.
  3224--3231, Oct. 2018.

\bibitem{wistort2008electric}
R.~{Wistort} and J.~R. {Smith}, ``Electric field servoing for robotic
  manipulation,'' in \emph{2008 IEEE/RSJ International Conference on
  Intelligent Robots and Systems}, Sept. 2008, pp. 494--499.

\bibitem{escaida20146D}
S.~{Escaida Navarro}, M.~{Schonert}, B.~{Hein}, and H.~W\"{o}rn, ``6d proximity
  servoing for preshaping and haptic exploration using capacitive tactile
  proximity sensors,'' in \emph{2014 IEEE/RSJ International Conference on
  Intelligent Robots and Systems}, Sept. 2014, pp. 7--14.

\bibitem{jiang2012seashell}
{L.-T. Jiang} and J.~R. {Smith}, ``Seashell effect pretouch sensing for robotic
  grasping,'' in \emph{2012 IEEE International Conference on Robotics and
  Automation}, May 2012, pp. 2851--2858.

\bibitem{gerratt2014stretchable}
A.~P. Gerratt, N.~Sommer, S.~P. Lacour, and A.~Billard, ``Stretchable
  capacitive tactile skin on humanoid robot fingers-first experiments and
  results,'' in \emph{2014 IEEE-RAS International Conference on Humanoid
  Robots}, 2014, pp. 238--245.

\bibitem{8830489}
A.~H. Memar and E.~T. Esfahani, ``A robot gripper with variable stiffness
  actuation for enhancing collision safety,'' \emph{IEEE Transactions on
  Industrial Electronics}, vol.~67, no.~8, pp. 6607--6616, 2020.

\bibitem{hirose-san}
T.~Hirose, Y.~Kakiuchi, K.~Okada, and M.~Inaba, ``Design of soft flexible
  wire-driven finger mechanism for contact pressure distribution,'' in
  \emph{2019 IEEE/RSJ International Conference on Intelligent Robots and
  Systems}, 2019, pp. 4699--4705.

\bibitem{8594159}
S.~Funabashi, S.~Morikuni, A.~Geier, A.~Schmitz, S.~Ogasa, T.~P. Torno,
  S.~Somlor, and S.~Sugano, ``Object recognition through active sensing using a
  multi-fingered robot hand with 3d tactile sensors,'' in \emph{2018 IEEE/RSJ
  International Conference on Intelligent Robots and Systems}, 2018, pp.
  2589--2595.

\bibitem{3DArchtect:Debkalpa:AFM2019}
D.~Goswami, S.~Liu, A.~Pal, L.~G. Silva, and R.~V. Martinez, ``3d-architected
  soft machines with topologically encoded motion,'' \emph{Advanced Functional
  Materials}, vol.~29, no.~24, p. 1808713, 2019.

\bibitem{kuo2020developing}
C.-C. Kuo, H.-Y. Kung, H.-C. Wu, and M.-J. Wang, ``Developing a hand sizing
  system for a hand exoskeleton device based on the kansei engineering
  method,'' \emph{Journal of Ambient Intelligence and Humanized Computing}, pp.
  1--13, 2020.

\end{thebibliography}
}

\end{document}